\documentclass[sigconf]{acmart}
\usepackage{xcolor}
\usepackage{caption}
\usepackage{subcaption}
\usepackage{array}
\usepackage{multirow}
\usepackage[ruled,vlined]{algorithm2e}
\usepackage{algorithmic}
\AtBeginDocument{%
  \providecommand\BibTeX{{%
    \normalfont B\kern-0.5em{\scshape i\kern-0.25em b}\kern-0.8em\TeX}}}

\acmConference[Sigspatial '21]{Sigspatial '21: ACM SIGSPATIAL}{November 02--05, 2021}{Sigspatial, Beijing}
\acmBooktitle{Sigspatial '18: International Conference on Advances in Geographic Information Systems, November 02--05, 2021, Beijing, China}
\acmPrice{15.00}
\acmISBN{978-1-4503-XXXX-X/18/06}

\begin{document}
\title{Towards Comparative Physical Interpretation of Spatial Variability Aware Neural Networks: A Summary of Results}

\author{Jayant Gupta}
\email{gupta423@umn.edu}
\affiliation{%
  \institution{University of Minnesota, Twin-Cities}
  \state{Minnesota}
  \country{USA}
}

\author{Carl Molnar}
\email{molna018@umn.edu}
\affiliation{%
  \institution{University of Minnesota, Twin-Cities}
  \state{Minnesota}
  \country{USA}
}

\author{Gaoxiang Luo}
\email{luo00042@umn.edu}
\affiliation{%
  \institution{University of Minnesota, Twin-Cities}
  \state{Minnesota}
  \country{USA}
}

\author{Joe Knight}
\email{jknight@umn.edu}
\affiliation{%
  \institution{University of Minnesota, Twin-Cities}
  \state{Minnesota}
  \country{USA}
}

\author{Shashi Shekhar}
\email{shekhar@umn.edu}
\affiliation{%
  \institution{University of Minnesota, Twin-Cities}
  \state{Minnesota}
  \country{USA}
}

\renewcommand{\shortauthors}{J. Gupta, C. Molnar, G. Luo, J. Knight, S. Shekhar}

\begin{abstract}
Given Spatial Variability Aware Neural Networks (SVANNs), the goal is to investigate mathematical (or computational) models for comparative physical interpretation towards their transparency (e.g., simulatibility, decomposability and algorithmic transparency). This problem is important due to important use-cases such as reusability, debugging, and explainability to a jury in a court of law. Challenges include a large number of model parameters, vacuous bounds on generalization performance of neural networks, risk of overfitting, sensitivity to noise, etc., which all detract from the ability to interpret the models. Related work on either model-specific or model-agnostic post-hoc interpretation is limited due to a lack of consideration of physical constraints (e.g., mass balance) and properties (e.g., second law of geography). This work investigates physical interpretation of SVANNs using novel comparative approaches based on geographically heterogeneous features. The proposed approach on feature-based physical interpretation is evaluated using a case-study on wetland mapping. The proposed physical interpretation improves the transparency of SVANN models and the analytical results highlight the trade-off between model transparency and model performance (e.g., F1-score). We also describe an interpretation based on geographically heterogeneous processes modeled as partial differential equations (PDEs).
\end{abstract}



\maketitle

\section{Introduction}

\begin{figure}[htp]
  \centering
    \includegraphics[width=0.95\linewidth]{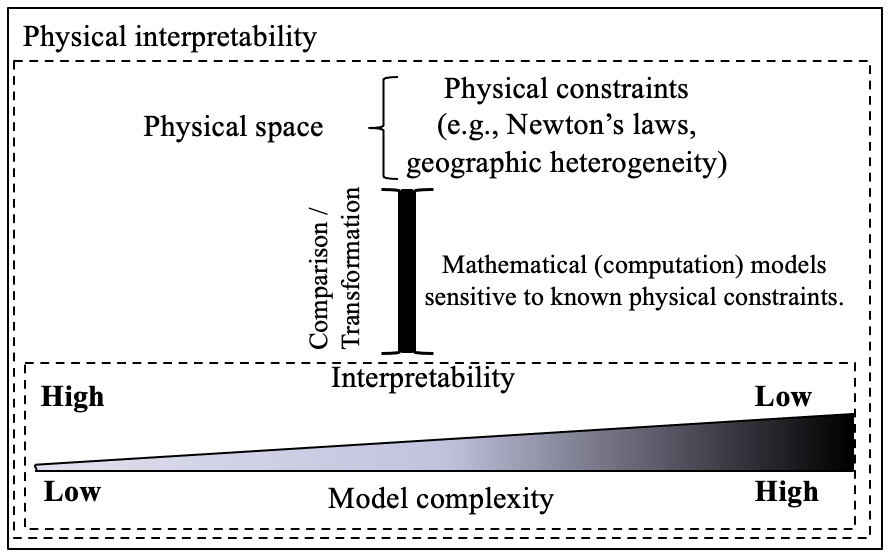}
  \caption{Towards physical interpretability.}
  \label{fig:phy_int}
\end{figure}

Interpretability of models changes with their complexity as simpler models (e.g., rule-based) are easier to interpret than complex models (e.g., multi-layer neural networks). For a model to be interpretable, it needs to be \textit{simulatable} i.e., it can be thought through by a human, \textit{decomposable} i.e., the input, parameters and output could be explained individually, and \textit{algorithmically transparent} i.e., the process to compute the output should be clear \cite{arrieta2020explainable}. However, this may not be enough for computational models built to represent physical processes which must adhere to different physical constraints (e.g., Newton’s laws, geographic heterogeneity, etc.). Figure \ref{fig:phy_int} illustrates the relation between physical interpretability and interpretability where comparison (i.e., comparative physical interpretation) or transformation (i.e., transformative physical interpretation) of a given model to other models or representations which are sensitive to known physical constraints and can serve as a bridge between the two. In this work, we strive towards comparison-based physical interpretability of spatial variability aware neural networks (SVANNs) \cite{gupta2020towards} which are sensitive to geographic heterogeneity. 

The physical interpretability of a spatial variability aware model is important to ensure its re-usability, to allow debugging of model errors, and to provide output determinism (i.e., enhance model explainability to a jury in a court of law). The need for interpretability can vary depending on the needs of the end-user \cite{rossi2019}. For example, artificial intelligence (AI) system builders need model interpretability to enhance model understanding and improve system performance, whereas lawmakers and medical practitioners need interpretability to enhance trust and confidence in the outputs. Regulatory bodies want model interpretability to ensure fairness for their constituents, while the affected users want to understand the factors responsible for an outcome.
\begin{table}[htp]
\small
\caption{User and applications of physical interpretability.}
\label{tab:user_app}
\begin{tabular}{|p{2.25cm}|p{5.25cm}|}
\hline
\textbf{User} & \textbf{Applications} \\
\hline
AI system builders & Enhancement of model understanding, Improvement in system performance.\\
\hline 
Lawmakers, medical practitioners & Enhancement of trust and confidence.\\
\hline
Regulatory bodies & Ensure trust and confidence in the system.\\
\hline 
Affected users & Understand factors responsible for an outcome.\\
\hline
\end{tabular}
\end{table}
The physical interpretation of a multi-layer neural network (MLNN) is hard due to the following factors. 

There are many reasons it is hard to interpret models built using multi-layer neural networks (MLNNs) or convolutional neural networks (CNNs) \cite{lecun1995convolutional} in physical terms. First, MLNNs have a large number of parameters which change with every learning step at different rates. Second, the networks can be highly unstable and small perturbations in the input (e.g., noise) and hyper-parameters can produce substantially different results \cite{goodfellow2014explaining}. Third, generalization error bounds are often vacuous with little explanatory power because they often show trends contradictory to the true error \cite{valle2020generalization}. Finally, the networks are at risk of over-fitting and may not generalize well on unseen data, leading to a strong bias in the system.

Related work in this area includes post-hoc interpretation of MLNNs using model-agnostic or model-specific techniques. Model-agnostic techniques can be used for any model and rely on techniques that simplify the model to a locally linear model (e.g., Local Interpretable Model-Agnostic Explanations (LIME) \cite{ribeiro2016should}); feature relevance estimation, which calculates desirable properties to predict a given class (e.g., Shapely Additive exPlanations (SHAP) \cite{lundberg2017unified}); and techniques to visualize input and output results. On the other hand, model-specific techniques for MLNNs split the inputs for each classification decision to highlight the contribution of its input elements \cite{montavon2017explaining}. Interpretability techniques for MLNNs can differ from those for CNNs, which generally: either explain the effect of an input on the network output, or interpret how each layer sees the input (i.e., what is learnt by each layer). For the first category, Matthew et al. \cite{zeiler2011adaptive} output feature maps for each layer showing strong and soft activation when an input image feeds-forward through a CNN. For the second category, a Layer Relevance Propagation technique \cite{bach2015pixel} was proposed, which generates a heatmap representing the contribution of a single pixel to the final prediction. One limitation of all these methods, however, whether model-agnostic or model-specific post-hoc methods, is that they do not consider physical constraints and properties of geographic spaces.

{\color{black}This paper investigates a novel approach to model interpretation based on geographically heterogeneous features. The features are constructed using well-known remote sensing indices. Remote sensing indices are derived from spectral imagery bands and owing to their simplicity many (indices) have a high level of interpretability for a given geographic region of study. Later, we also discuss an interpretation based on geographically heterogeneous processes where the processes are represented using partial differential equations (PDEs). PDEs are fundamental scientific elements which are used to understand many real-world phenomena (e.g., sound, heat, etc.) which means they are also highly interpretable. The two (i.e., indices and PDEs) help in developing an independent physical interpretations of SVANN based models.} 

We find that our spatial heterogeneity feature based interpretation of SVANNs improves their decomposability which are further substantiated by a case study on wetland mapping using remote-sensing indices. 

Theoretically, our process-based interpretation of SVANNs should improves their simulatibility, decomposability, and algorithmic transparency, however, this is yet to be validated.

Following is the summary of our key contributions:

\textbf{Contributions:}
\begin{itemize}
    \vspace{0pt}
    \item We provide two physical interpretations of SVANNs as follows:
    \begin{itemize}
        \item Geographically heterogeneous feature-based using remote sensing indices.
        \item Geographically heterogeneous process-based using partial differential equations (PDEs).
    \end{itemize}
    \item We provide a case study on wetland-mapping to validate the physical interpretation based on geographically heterogeneous features.
    \begin{itemize}
        \item Investigate image upsampling for multi-scale data fusion preprocessing to improve model results.
    \end{itemize}
    \vspace{0pt}
\end{itemize}

\textbf{Scope:} In this paper we limit our discussion of geographically heterogeneous process-based interpretation to PINNs \cite{raissi2019physics}, which is one of the many techniques to solve partial differential equations (PDEs) using multi-layer neural networks (MLNNs). Further, experimental evaluation of process-based interpretation is not explored in this paper and is planned in the near future. The experimental evaluation of Geographically heterogeneous feature-based interpretation is limited to two remote sensing indices: the Normalized Difference Vegetation Index (NDVI) \cite{TUCKER1979127} and the Normalized Difference Water Index (NDWI) \cite{mcfeeters1996}, and two case-study regions, which characterize their uses. Further, our approximation of rule-based methods is kept simple (i.e., based on only one index) specifically to serve as a highly interpretable baseline method.
\\\textbf{Organization:} The paper is organized in the following manner: Section \ref{sec:RWB} gives a brief background on spatial heterogeneity and SVANNs. Section \ref{sec:approach} describes our geographically heterogeneous feature-based approach to interpreting SVANNs. Section \ref{sec:eval} describes the wetland-mapping based evaluation of the feature-based interpretation and performance based evaluation. Section \ref{sec:results} discusses the experimental results and describe process-based model interpretability using PDEs. Finally, we conclude and consider future work in Section \ref{sec:CFW}.
\section{Related Work and Background}
\label{sec:RWB}

\subsection{Spatial Heterogeneity}
\label{spatial_heterogeneity}

\begin{figure}[htp]
  \centering
    \includegraphics[width=0.99\linewidth]{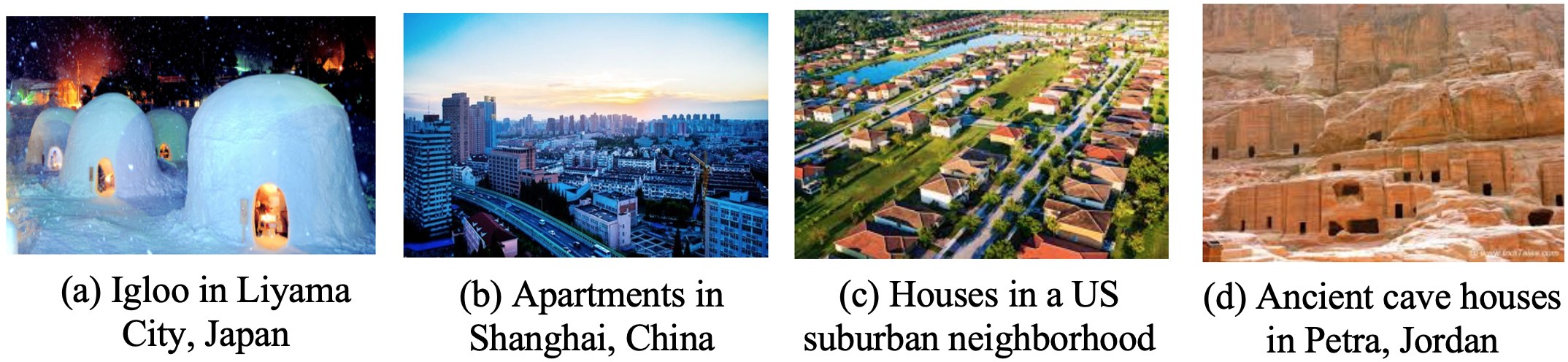}
  \caption{Spatial variability in houses and their surroundings.}
  \label{fig:spatial_variability}
\end{figure}

Geographic properties differ across different areas giving rise to varied geophysical and cultural phenomena. This spatial variability results in different physical models across geographic areas. Knowledge of spatial variability is necessary to understand the spatial patterns of events and objects over an area that vary spatially \cite{turner2005causes}. A computational model can be affected by two types of spatial variability: variability in the object of interest itself, which may differ in shape, size, or both; and variability in the background of the object of interest. For example, a computational model that is trained to find residential housing in the the US may have difficulty finding houses in other places where housing construction is adapted to different local climates or other conditions (e.g., cave houses in Petra, igloos in polar regions, etc.). Here the neighbor surroundings differ as well. Figure \ref{fig:spatial_variability} shows the spatial variability in houses and their background across the globe.

Spatial variability has been observed in many geo-phenomena including climatic zones, USDA plant hardiness zones \cite{USDA_hardiness}, and terrestrial habitat types (e.g., forest, grasslands, wetlands, and deserts). The differences in climatic zones affect the plant and animal life of the region. Spatial variability is considered as the second law of geography \cite{leitner2018laws} and has been adopted in regression models (e.g., Geographically Weighted Regression \cite{brunsdon1998geographically}) to quantify spatial variability, the relationships among variables across a study area.

\begin{figure}[htp]
  \centering
    \includegraphics[width=0.95\linewidth]{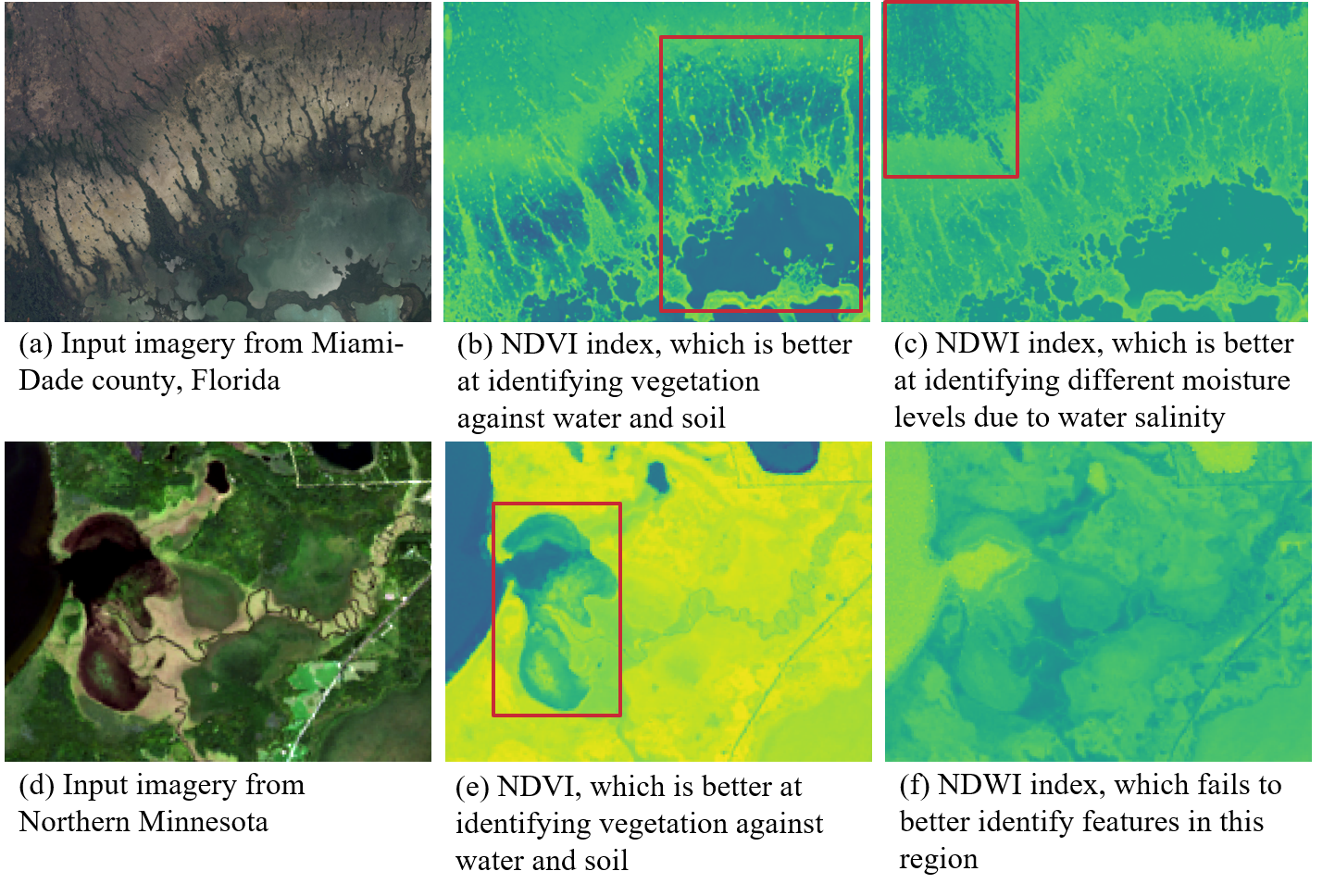}
  \caption{Using the same remote sensing indices across spatially heterogeneous regions.}
  \label{fig:indices_comparison}
\end{figure}

\textbf{Spatially heterogeneous features:} Remote sensing indices are becoming increasingly used for identifying geophysical features such as plants, soil, and water \cite{gaitan2013indices}. Typically the index would be used as part of subsequent analysis steps; they are indicators of information, not results by themselves. These indices also have limitations in the extent of the area which can be evaluated using them, due to spatial heterogeneity between regions \cite{gaitan2013indices}. For example, some coastal  regions (e.g., Miami-Dade County, Florida) consist of  freshwater and intertidal regions in close proximity, which causes high variation in vegetation density and moisture levels. In this case, it can be beneficial to use a remote sensing index that measures differences in vegetation density, such as NDVI, or an index which specializes in measuring differences in vegetation moisture, such as NDMI. By selecting either of these remote sensing indices in this region, we can see an increase in the ability to delineate wetland features (Figure \ref{fig:indices_comparison}a-c).

However, for regions which consist of only dense freshwater forested vegetation, we may wish to select different remote sensing indices for identifying features. While NDVI may still contribute some useful information in such a region, the need for NDMI is diminished, due to a similar moisture level in the region's vegetation (Figure \ref{fig:indices_comparison}d-f).

Current multi-layer neural network and convolution neural network based models follow a One-Size-Fits-All (OSFA) where models are trained without consideration to spatial variability. To address the issue, we proposed spatial variability-aware neural networks (SVANN) \cite{gupta2020towards} whose network weights and architecture can differ across locations to account for the heterogeneity in a geographic space that follows Newton's laws. The approach was evaluated using CNN based deep neural networks. 

\subsection{Spatial Variability Aware Neural Networks}
There are two types of SVANN approaches as described below.
\\\textbf{SVANN-I} \cite{gupta2020towards} is a spatially explicit neural network model where each parameter (e.g., weight) is a function of the model's location $loc_M$. The architecture $f$ is composed of a sequence of $K$ weight functions or layers $(w^1(loc_M),...,w^K(loc_M))$ which map a geographic location-based training sample $x(loc_S)$ to a geographic location-dependent output $y(loc_M)$ as follows: 
\begin{align}
\label{eq4}
y(loc_M) &= f(x(loc_S) ; w^{1}(loc_M),...,w^{K}(loc_M)), 
\end{align}
where $w^i(loc_M)$ is the weight vector for the $i^{th}$ layer. In this approach we assume that the architecture is location invariant i.e., $K$ is constant for all the models.

Figure \ref{Figure:WM_IO} shows an example of SVANN-I (termed as SVANN) approach where the geographical space has 4 zones and deep learning models are trained for each zone separately. For prediction, each zonal model predicts the test samples in its zone. 
\\\textbf{SVANN-E} \cite{gupta2020TIST} is a spatially explicit neural network model whose architecture is a function of model location $loc_M$. The architecture $f_{loc_M}$ is a sequence of $K$ weight functions or layers mapping a geographical location based training sample $x(loc_S)$ to a geographical location dependent output $y(loc_M)$ as follows. 

\begin{align}
\label{eq_svann_e}
y(loc_M) &= f_{loc_M}(x(loc_S) ; w^{1}(loc_M),...,w^{K}(loc_M)), 
\end{align}
where $w^i(loc_M)$ is the weight vector for the $i^{th}$ layer. Further, as the architecture varies across locations, it implies that the weights are also a function of location. 

SVANN-I is a special case of SVANN-E; therefore, the classification accuracy of SVANN-I can never be greater than the classification accuracy of SVANN-E \cite{gupta2020TIST}.

In the following text, we will use SVANN to refer to SVANN-I models. In case otherwise, we will use SVANN-E.




\section{Proposed Approach}
\label{sec:approach}
Our approach is to interpret SVANNs using geographically heterogeneous features. In this section, we define what we mean by simulatibility, decomposability, and algorithmic transparency (Section \ref{subsec:base_def}). Next, we describe an interpretation of SVANNs using geographically heterogeneous features and compare the interpretability of different models.
\subsection{Basic Definitions}
\label{subsec:base_def}
\textit{Simulatability} denotes the ability of a model to be simulated, or thought about strictly, by a human. Hence, model complexity (e.g., number of parameters, architecture) is relevant, where the model needs to be self-contained enough for a human to think and reason about it as a whole. For example, NDVI can be easily simulated given a set of Red and NIR bands (Equation \ref{eq: NDVI}), in contrast, a decision tree with a depth of $100$ is hard to simulate.
\\\textit{Decomposability} represents the ability to explain each part of a model (input, parameter and calculation). Decomposability requires every input to be readily interpretable and every part of the model to be understandable by a human without the need for additional tools. For example, a linear regression model is decomposable as each parameter is associated with a meaningful variable, whereas the parameters in deep learning models are hard to interpret as the models rely on their machine based representation learning.
\\\textit{Algorithmic transparency} deals with the ability of a user to understand the process followed by the model to produce any given output from its input data. For example, a linear model is considered transparent because its error surface can be readily understood. In contrast, deep neural network architectures are considered opaque as the loss landscape cannot be fully observed, and instead has to be approximated through heuristic optimization (e.g., stochastic gradient descent). Algorithmically transparent models are constrained by means of mathematical analysis and methods \cite{arrieta2020explainable}.

{\color{black}
\subsection{Feature-Based Interpretation}
A \textit{transformation-based approach} converts the given model (say model 1) to a physically interpretable model. For physically interpretable models, the mapping is relatively simpler. For example, if we consider a partial differential equation as model 1 and a set of rules as model 2. Then, the domain space of model 1 can be mapped to model 2 depending on the physical constraints. On the other hand, model transformation can be difficult for complex models or models which do not adhere to physical constraints. For such models we describe the comparison-based approach which is also the focus of this work. 

A \textit{comparison-based approach} compares the output of the given model to a physically interpretable model. For example, consider a model (say Model 1) which has 4 parameters (or weights) out of which 3 have known physical properties, however, 1 does not. Further, we have another Model (say Model 2) which can perform the same task and is physically interpretable. Then, the model may be useful to identify the parameter with the unknown properties. Figure \ref{fig:Approach} below shows the contrast between the two approaches.
\begin{figure}[htp]
\vspace{-5pt}
  \centering
    \includegraphics[width=0.95\linewidth]{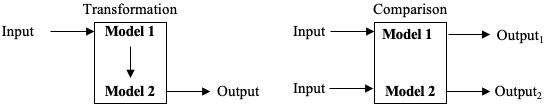}
  \caption{Physical interpretability by Transformation or Comparison. We assume that Model $2$ is a physically interpretable model.}
  \label{fig:Approach}
\end{figure}

A more general form of this example would be a model with $n$ output parameters where $d$ features have known physical properties whereas $n-d$ parameters have unknown physical properties and the physical interpretability of these features is established through a more physically interpretable model.

To show how spatial heterogeneity affects model interpretability, we use the example of two spatially heterogeneous locations, $R_1$ and $R_2$, where $R_1$ is a region with a high density of forests and $R_2$ is a region with a significant water and wetland coverage. We also have two types of spatially heterogeneous features (e.g., remote sensing indices), one of which is useful to delineate between different vegetation types (say $RSI_1$) and one that delineates water-based land-cover features ( say $RSI_2$). 

A SVANN-based approach will consider training two distinct models to classify land cover for each region. The SVANN outputs are locally dependent weights and lack any physical interpretation. Now, consider two rule-sets built using each of the two heterogeneous features ($RSI_1$ and $RSI_2$) to classify the land cover across $R_1$ and $R_2$. 


A comparison of output weights from SVANN with the rule-sets would in theory highlight the set of weights responsible for the two spatially heterogeneous regions (i.e., $R_1$ and $R_2$). Furthermore, the prediction of spatially heterogeneous weights should behave similar to the spatially heterogeneous features for any input. For example, the set of SVANN weights which behave similar to rule-sets for $RSI_1$ are suitable for forest-based regions and similarly the other set for water and wetland coverage. In this work, we use accuracy and F1 measures as a proxy to compare the model behavior.

The example above helps to separate the weights, which are otherwise indistinguishable based on their sensitivity to heterogeneity. This clearly shows that comparison-based approach helps to improve the decomposability of the SVANN weights.

The above interpretation is highly useful when the nature of channels to SVANN based models is not known. In such instances, one can attempt to compare various SVANN models with simple rule based approach to identify the channel used for a given SVANN model.

Comparative physical interpretation depends upon the complexity of the physically interpretable models. For example, models such as partial differential equations (PDEs) have a higher representative power than rule-based models. Furthrmore, PDEs are known to be far more physically interpretable due to their extensive usage to model physical processes (e.g., heat flow). We briefly describe an approach of modeling SVANNs in Section \ref{subsec:PDE-based}.
}
\section{Evaluation Procedure}
\label{sec:eval}
This section describes how we evaluated an interpretation of SVANNs based on a wetland mapping case-study for selecting remote sensing indices based on geographic location. We present the problem statement, some background on the remote sensing indices used in the case-study, and experiment design. We then describe the model architecture, evaluation metrics, and the dataset used in the case-study. Results of the evaluation are given in Section \ref{sec:results}.
\subsection{Problem Statement}
In this case-study, our goal was to use a SVANN approach to aid in the selection of remote sensing indices for a set of wetland mapping tasks. A wetland refers to a flooded area of land having a distinct ecosystem based on hydrology, hydric soils, and vegetation adapted for life in water-saturated soils \cite{keddy2010wetland}. Wetland inventory maps are essential for wetland management, protection and restoration. However, development of highly accurate wetland inventories can be expensive and technically challenging. Further, they require periodic updates due to seasonal changes, land use change and climate change. In recent work, semi automated wetland mapping approaches are common (e.g., \cite{kloiber2015semi}), which use a combination of image segmentation and random forest classification, along with aerial photo interpretation. However, these approaches are iterative and time-consuming to derive, and heavily require expert analysis.

\begin{figure}[htp]
\vspace{-5pt}
  \centering
     \includegraphics[width=1\linewidth]{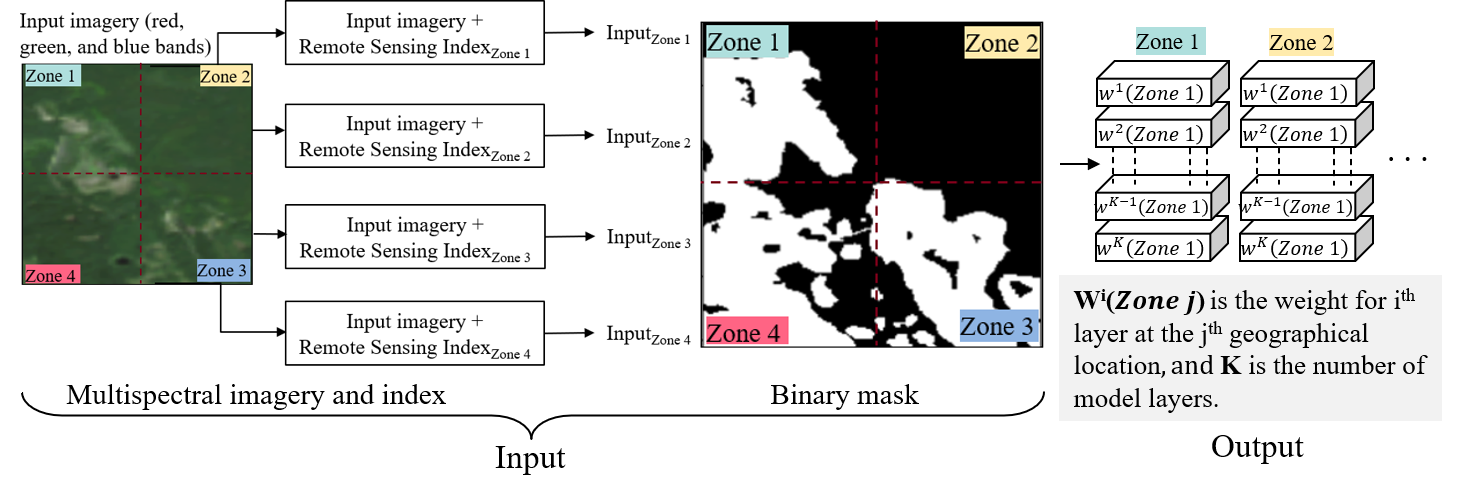}
  \caption{Example input and output for the SVANN evaluation.}
  \label{Figure:WM_IO}
\end{figure}

We defined our wetland mapping task as a pixel-level binary image segmentation process where we label each pixel of an image as a class, consisting of wetland or non-wetland. Figure \ref{Figure:WM_IO} shows the input and output of the system: the three color (i.e., red, green, and blue) channels, acquired from multispectral satellite imagery, joined with an index channel (e.g., NDVI). Our input also consisted of the corresponding binary wetland mask as a true label. The white region in the binary mask represents the pixels that belong to a wetland class. The two inputs are used to train the final semantic segmentation models.
Various remote sensing indices are used in the domain of wetland mapping. These indices are optimized for identifying different land cover features. Remote sensing indices must be carefully chosen for a specific geographic location, due to differences in climate and ecosystem. In this case-study, we do so by comparing semantic segmentation results between CNN models which use different remote sensing indices across spatially heterogeneous locations.

\subsection{Geographically Heterogeneous Indices}
\label{subsec:indices}
Each multispectral Landsat-8 image in the study area consisted of several bands (e.g., Blue, Green, Red, Near Infrared (NIR), Shortwave Infrared 1). Each band represents a particular portion of the electromagnetic spectrum or radiation. These bands can be used to calculate remote sensing indices to better delineate land-cover features in a geographical area. We chose to perform our evaluations using two well-known remote sensing indices which are relevant to wetlands in northern Minnesota. 

The first index was the Normalized Difference Vegetation Index (NDVI) \cite{TUCKER1979127}. It is formulated as Equation \ref{eq: NDVI} by Near-Infrared (NIR) and visible red spectrum (Red), which corresponds to band 5 and band 4 of the Landset-8 dataset. The formula for NDVI is given as:
\begin{equation}
    \text{NDVI}= \frac{\text{NIR - Red}}{\text{NIR + Red}}
    \label{eq: NDVI}
\end{equation}

The second index was the Normalized Difference Water Index (NDWI) \cite{mcfeeters1996} formulated as Equation \ref{eq: NDWI}, where band 3 of the Landset-8 dataset represents the visible green spectrum (Green). The formula for NDWI is given as:
\begin{equation}
    \text{NDWI}= \frac{\text{Green - NIR}}{\text{Green + NIR}}
    \label{eq: NDWI}
\end{equation}

While both NDVI and NDWI have a range of $\lbrack -1, 1 \rbrack$, NDVI is more sensitive to chlorophyll content in leaves \cite{NASSERMOHAMEDEID202066}, so it has been found useful for distinguishing vegetated and non-vegetated areas \cite{jknight2013}. In contrast, NDWI is more sensitive to water content, and a positive NDWI value indicates water areas and a negative value indicates a non-water surface \cite{rs11232834}.

\subsection{Experiment Design}
 To evaluate the efficacy of SVANN as a method of selecting remote sensing indices based on location, we first trained four SVANN models (SVANN Models 1-4), where SVANN Models 1 and 2 were trained and evaluated on a Freshwater Forested/Shrub Wetland region, and SVANN Model 3 and 4 were trained and evaluated on a Lakes Wetland region. SVANN then selects the best performing model for each region. 
 
\textit{SVANN Versus Rule-Based OSFA:} To compare with a rough simulation of traditional wetland mapping automation, we built two baseline rule-based OSFA classifiers (Rule-Based Models 1 and 2), which use an established ruleset for each remote sensing index to classify wetland features. 
 
\begin{figure}[htp]
 \vspace{-5pt}
  \centering
  \includegraphics[width=1\linewidth]{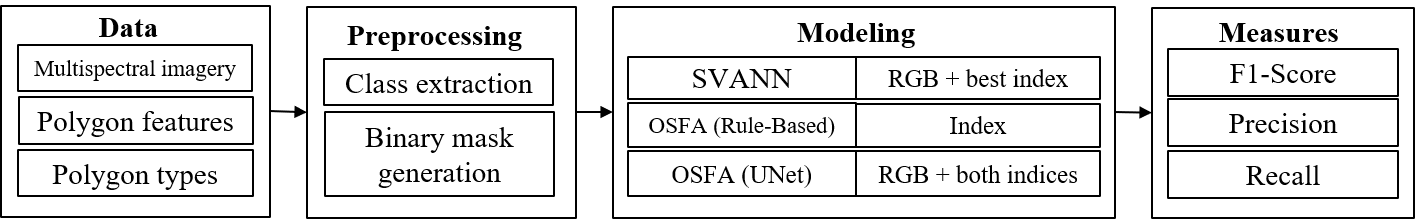}
  \caption{Experiment design used in the evaluation.}
  \label{fig:experiment_flow}
\end{figure}

\textit{SVANN Versus UNet OSFA:} To compare with a standard deep learning approach, which does not consider spatial variability when selecting input features, we trained an OSFA UNet model (OSFA Model), which used both remote sensing indices and both geographic regions as its input. We then compared the best performing SVANN models against the OSFA model on the same region. Figure \ref{fig:experiment_flow} shows the experiment design used in the evaluation.

\textit{Comparative physical interpretation:} To validate the comparative physical interpretation of SVANNs we compare their F1-score to the rule-based models for each test area. If the behavior of a SVANN model is like a particular rule-based approach, it suggests that the SVANN weights are suitable for that area over the other. 

The validation becomes further relevant as the zonal boundary starts to overlap i.e., the test area starts to overlap. We plan to explore the validation on overlapping test regions in the future work.


 \begin{figure}[htp]
 \vspace{-10pt}
  \centering
  \includegraphics[width=1\linewidth]{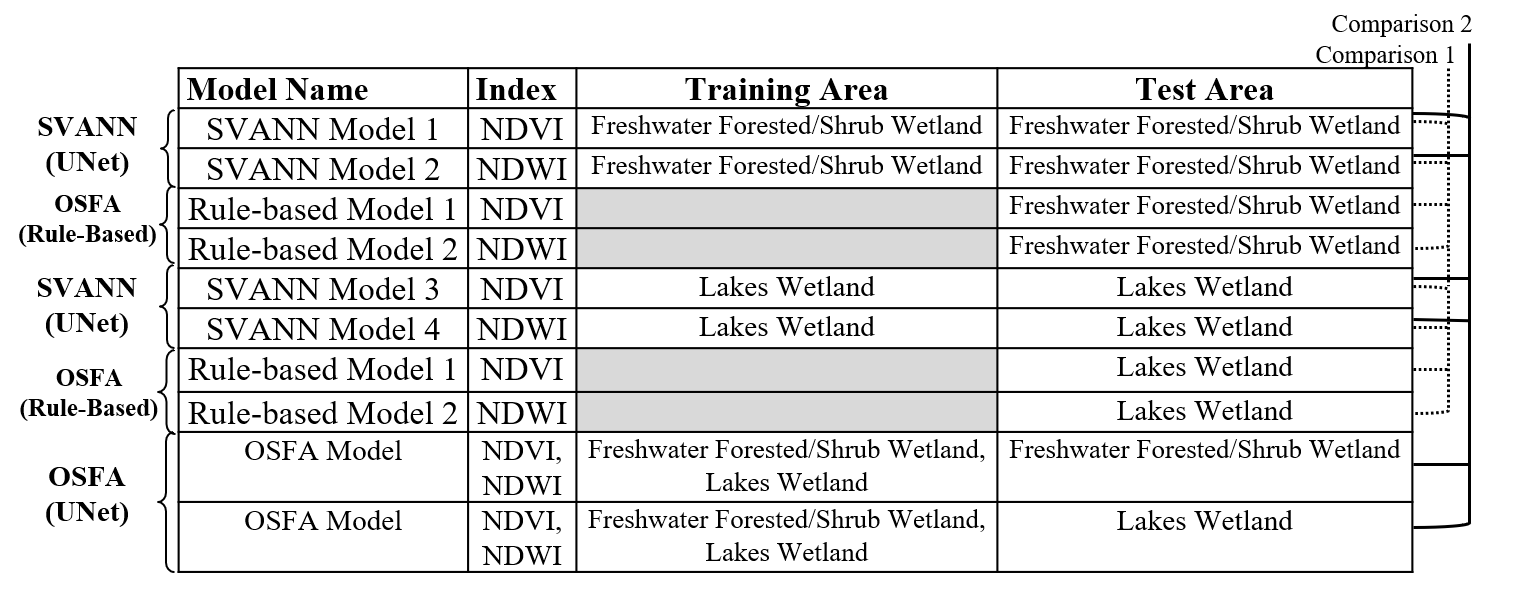}
  \caption{Experiment design showing the comparisons made in the case-study to distinguish SVANN as a method of feature selection.}
  \label{fig:experiment_design}
\end{figure}

\subsection{Modeling}
\textbf{Rule-Based Methodology:}
We built baseline rule-based classifiers for wetland classification using the NDVI and NDWI remote sensing indices.

NDVI is frequently used as a tool for wetland identification. We defined the rules for our rule-based NDVI index wetland classification following established value ranges for certain wetland and non-wetland features (e.g., \cite{brown2015ndvi}).
\begin{table}[htp]
\small
\caption{NDVI Rule-Based Classification}
\label{tab:NDVI_Rules}
\begin{tabular}{|p{4.5cm}|p{2cm}|}
\hline
\textbf{Classification} & \textbf{Values} \\
\hline
Water & $-1 \text{ to } -0.1$\\
\hline 
Rocks, clouds, buildings, etc & $-0.1 \text{ to } 0.1$\\
\hline
Sparse Wetland Vegetation & $0.1 \text{ to } 0.73$\\
\hline 
Dense Non-Wetland Vegetation & $0.73 \text{ to } 1$\\
\hline
\end{tabular}
\end{table}

NDWI is commonly used for water identification, not vegetation. As a result, common value ranges for NDWI are derived specifically for identifying water features \cite{ji2009ndwithresholds}. Therefore, we manually defined the rules for our rule-based NDWI index wetland classification by adjusting the classification threshold to account for vegetation features as well as water features.
\begin{table}[htp]
\small
\caption{NDWI Rule-Based Classification}
\label{tab:NDWI_Rules}
\begin{tabular}{|p{4.5cm}|p{2cm}|}
\hline
\textbf{Classification} & \textbf{Values} \\
\hline
Non-Wetland & $-1 \text{ to } -0.6$\\
\hline 
Wetland & $-0.6 \text{ to } 1$\\
\hline
\end{tabular}
\end{table}

By thoroughly evaluating the generated prediction masks, we found these two rulesets to be fairly successful on our case-study regions. These rule-based approaches correctly classify wetland features with reasonable accuracy and make highly interpretable predictions. Therefore, we were able to use these rulesets in our rule-based approach as a robust, interpretable baseline to compare against our SVANN model.

\textbf{Semantic Segmentation Methodology:} We built the SVANN models using the UNet \cite{ronneberger2015u} architecture. UNet is an established technique for image segmentation in biomedical imagery that has been adapted for aerial imagery \cite{iglovikov2018ternausnet}. This architecture follows an encoder-decoder framework which combines local pixel information with its context. To achieve this, high resolution features from the contracting set of layers (i.e., context) are concatenated with the output from the up-sampled images (i.e., localization). We provide a brief overview of UNet in Appendix \ref{App:UNET}.

\subsection{Evaluation Metrics}
Since our case-study was a binary classification task, we were able to derive evaluation metrics by constructing a confusion matrix consisting of correct and incorrect model predictions, compared with the ground truth labels. The values in the confusion matrix were labeled as true positive (TP), true negative (TN), false positive (FP), and false negative (FN). Since our study had an imbalanced class distribution, it was necessary to select evaluation metrics which punish false predictions and therefore measure the relevance and completeness of the model predictions \cite{cloudunet2019}. For this reason, we used the metrics precision and recall, called respectively the complement of commission errors and the complement of emission errors by the remote sensing community. We also used F1-score which balances the trade-off between precision and recall. Detailed description of precision, recall, and F1-score can be found in Appendix \ref{App:PRF}.

\subsection{Dataset}
\begin{figure}[htp]
\vspace{-10pt}
  \centering
   \includegraphics[width=0.89\linewidth]{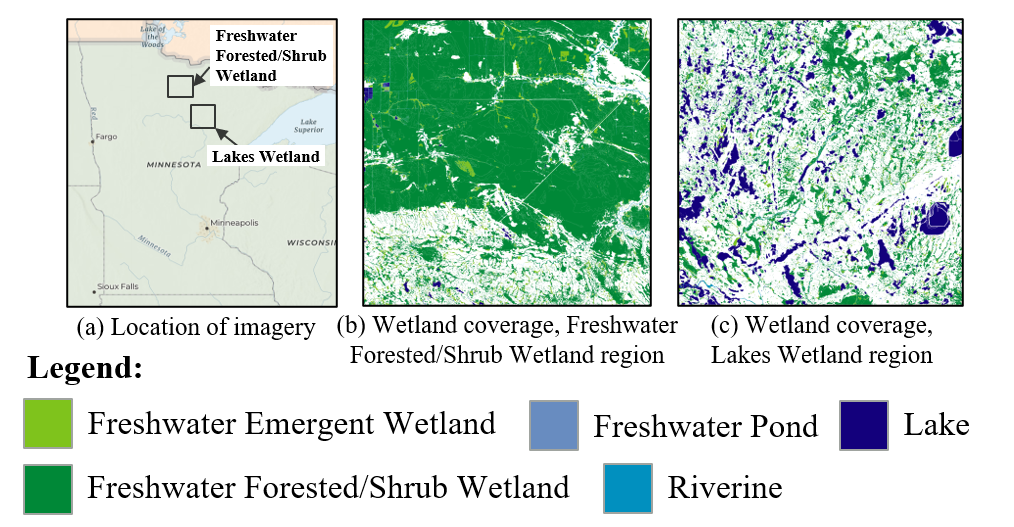}
   \caption{Location and types of wetland (Best in color).}
  \label{fig:WM}
\end{figure}

We used multispectral imagery from Landsat-8 Operational Land Imager (OLI) \cite{landsat2014} and part of a set of wetland maps developed by the National Wetlands Inventory (NWI) \cite{NWI} for the mapping task. The multispectral imagery consisted of the Blue, Green, Red, NIR, SWIR1, and SWIR2 bands, each with a spatial resolution of 30m and acquired on Sept. 11, 2020. The NWI wetland maps in this case-study consisted of 718,304 features, which represent the extent, approximate location, wetland types and surface water habitats in northern Minnesota. All the features in the NWI wetland map were classified according to the Cowardin classification system \cite{cowardin1979classification}. In this work, we limited our analysis to two study areas located in northern Minnesota. The first region predominantly consists of Freshwater Forested/Shrub wetlands. The second region mainly consists of small Lakes wetlands. Figure \ref{fig:WM} (a) shows the location of the two study areas. 
Overall, the study regions had 5 different types of wetlands, as shown in Figure \ref{fig:WM}b-c.
\vspace{-10pt}

\subsection{Preprocessing}
\label{subsec:preproc}
To align and limit the analysis to the study area, the wetland shapefile and multispectral imagery were first re-projected to the WGS84 reference system. Next, the wetland shapefile and imagery were cropped to the case-study regions. 

Since the Landsat-8 imagery has a spatial resolution of 30m \cite{landsat2014}, while the NWI wetland shapefile has a positional accuracy of about 2.2m \cite{minnesota_dnr_2019}, we then up-sampled the multispectral imagery by a factor of 4 using bilinear interpolation, which is a nearest-neighbor algorithm for increasing the number of pixels. Up-sampling provided a $4^2$, or $16$, times total increase in imagery pixels, resulting in $8,306 \times 5,434$ pixels from the Freshwater Forested/Shrub Wetland region and $9,046 \times 5,709$ pixels for the Lakes Wetland region. 

\begin{figure}[htp]
  \centering
   \includegraphics[width=0.99\linewidth]{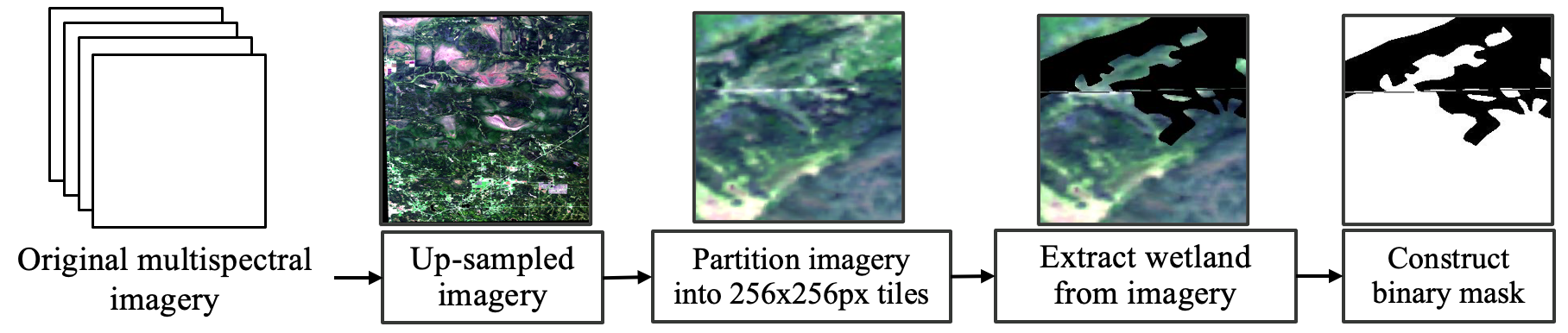}
   \caption{Preprocessing imagery and polygon based wetland maps.}
  \label{fig:PP}
\end{figure}

We then partitioned the imagery (Figure \ref{fig:PP}) and its mask into images having the dimensions of $256 \times 256$ pixels. These steps resulted in $672$ and $770$ samples for the Freshwater Forested/Shrub Wetland and Lakes Wetland regions respectively. Partitions at the right and bottom edges of the imagery were removed, as a majority of the area was empty.

The binary masks for training and testing were created in two steps. First, we used the polygons in the wetland shapefile to extract the wetland from the imagery for each region. We then used the extracted wetland imagery to create a binary mask for model training. Figure \ref{fig:PP} shows the initial imagery from the Freshwater Forested/Shrub Wetland region, an example of extracted wetland imagery for each image partition, and the corresponding binary mask where white pixels represent wetlands. The dataset was split into training ($\sim80\%$), validation ($\sim10\%$), and testing ($\sim10\%$) datasets to build, validate and test the models.

\section{Results and Discussion}
\label{sec:results}

In our first experiment, we evaluated the effect of training SVANN with and without up-sampled imagery. The second and third set of experiments compared trained SVANN models against two one-size-fits-all approaches, a rule-based approach (Rule-Based Models 1 and 2) and a deep learning UNet model (OSFA Model). Since our overarching goal is to interpret the SVANN approach, we also include a discussion of SVANN's interpretability.

\subsection{Up-Sampling Versus Non Up-Sampling}
In the preprocessing stage of the case-study, we compared our results when training SVANN models with up-sampled imagery (Up-Models 1-4 in Table \ref{tab:results_comparison}, where the model chosen by SVANN is shown in bold) versus training with non up-sampled imagery (Models 1-4). Previously, we had found that the lower spatial resolution of the Landsat-8 multispectral imagery (30m), compared with the positional accuracy of the NWI wetland shapefile (approx. 2.2m), caused a significant amount of fine-scale information to be lost when preprocessing the input data for the SVANN models, which made model training difficult for the Lakes Wetland region. As discussed earlier, we addressed this issue by first up-sampling the multispectral imagery using bilinear interpolation before generating the binary wetland masks. Then, we used the up-sampled imagery and binary wetland masks as model inputs for the up-sampled models (Up-Models 1-4). The input for the non up-sampled models (Models 1-4) was the original imagery, which was used to generate their binary wetland masks.

\begin{figure}[htp]
  \centering
   \includegraphics[width=0.89\linewidth]{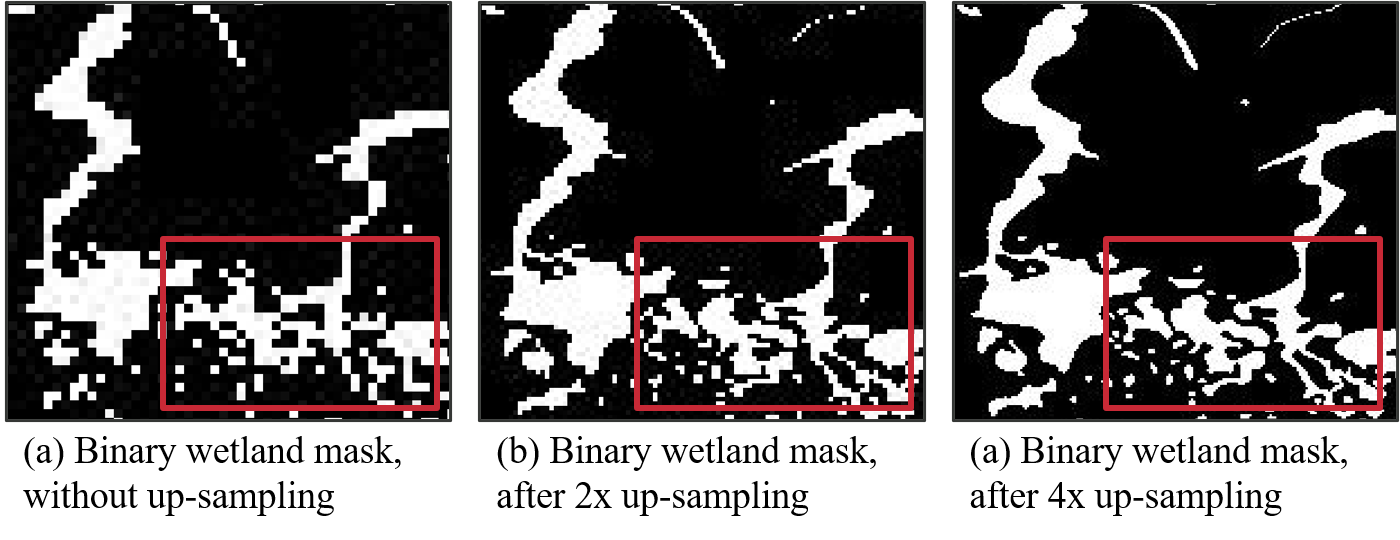}
   \caption{Imagery up-sampling process, demonstrating the increase of wetland information retained by the binary mask when the imagery is up-sampled.}
  \label{fig:upsample}
\end{figure}

Table \ref{tab:results_preprocessing} shows the results of this comparison. In the Lakes Wetland region, Up-Model 4 sees a significant improvement (+0.440) in validation accuracy and evaluation metrics (e.g., +0.298 in F1 score) over Model 4. Our results suggest that the up-sampling of the imagery provided a significant improvement in the ability of the model to classify wetlands in this region. This improvement is likely due to the prevalence of small wetland features (i.e., Figure \ref{fig:WM}c), which cannot be accurately represented by a low-resolution binary wetland mask. In the Freshwater Forested/Shrub Wetland region, Up-Model 1 sees a slight decrease in F1 score (-0.032), but an increase in validation accuracy (+0.149), over Model 1. We speculate that this is due to the homogeneity of the wetlands in the region, which allowed the low-resolution binary mask to capture the wetland information with reasonable accuracy, and made evaluation on the test data slightly more difficult. Overall, we found the up-sampling to be useful for two reasons: firstly, model performance was overall improved, since the up-sampled models produced more consistent and overall better classification accuracy. Secondly, model decomposability was improved, since the up-sampled models were trained using more detailed and contiguous binary wetland masks, which are more clearly understood by humans than the non up-sampled binary masks (see Figure \ref{fig:upsample}).

\begin{table*}
\caption{Experimental results showing the effect of up-sampling on model performance.}
\label{tab:results_preprocessing}
\begin{tabular}{|l l l l l l l l l l l|}
\hline
\textbf{Model} & \textbf{Index} & \textbf{TN} & \textbf{FP} & \textbf{FN} & \textbf{TP} & \textbf{Precision} & \textbf{Recall} & \textbf{F1} & \textbf{Val. Acc.} & \textbf{Epochs} \\
\hline
\multicolumn{11}{|c|}{Test Area: Freshwater Forested/Shrub Wetland region}\\
\textbf{Up-Model 1} & \textbf{NDVI} & \textbf{1,114,819} & \textbf{432,164} & \textbf{763,862} & \textbf{2,145,603} & \textbf{0.832} & \textbf{0.737} & \textbf{0.782} & \textbf{0.786} & \textbf{3}\\
Up-Model 2 & NDWI & 240,074 & 1,839,944 & 13,532 & 2,362,898 & 0.562 & 0.994 & 0.718 & 0.758 & 3\\
\hline 
\textbf{Model 1} & \textbf{NDVI} & \textbf{94,020} & \textbf{27,567} & \textbf{30,313} & \textbf{126,628} & \textbf{0.821} & \textbf{0.807} & \textbf{0.814} & \textbf{0.637} & \textbf{3}\\
Model 2 & NDWI & 87,923 & 33,664 & 57,227 & 99,714 & 0.748 & 0.635 & 0.687 & 0.594 & 3\\
\hline 
\hline
\multicolumn{11}{|c|}{Test Area: Lakes Wetland region}\\
Up-Model 3 & NDVI & 2,266,152 & 847,533 & 581,179 & 1,351,408 & 0.615 & 0.699 & 0.654 & 0.696 & 3\\
\textbf{Up-Model 4} & \textbf{NDWI} & \textbf{2,423,985} & \textbf{689,700} & \textbf{455,228} & \textbf{1,477,359} & \textbf{0.682} & \textbf{0.764} & \textbf{0.721} & \textbf{0.723} & \textbf{3}\\
\hline 
Model 3 & NDVI & 222,181 & 63,466 & 11,497 & 18,248 & 0.223 & 0.613 & 0.327 & 0.231 & 3\\
\textbf{Model 4} & \textbf{NDWI} & \textbf{237,436} & \textbf{48,211} & \textbf{8,830} & \textbf{20,915} & \textbf{0.303} & \textbf{0.703} & \textbf{0.423} & \textbf{0.283} & \textbf{3}\\
\hline
\end{tabular}
\end{table*}

\subsection{SVANN Versus One-Size-Fits-All}
We then compared SVANN against the two different OSFA approaches. Table \ref{tab:results_comparison} shows the results of the experiments.

\subsubsection{SVANN Versus Rule-Based OSFA}
To compare the SVANN models with the rule-based OSFA approach, we made two sets of regional comparisons: In the Freshwater Forested/Shrub Wetland region, we compared SVANN Model 1 and Rule-based Models 1 and 2. In the Lakes Wetland region, we compared SVANN Model 4 and Rule-based Models 1 and 2. For both of these regions, we found that the SVANN model outperformed the rule-based approaches on all three evaluation metrics (i.e., precision, recall, and F1 score). 

This comparison tells us two things: first, it suggests that deep learning is superior to the baseline rule-based approach, since the deep learning models in the case-study are capable of learning more complex features of wetland delineation than the rule-based approach. Second, it suggests that SVANN is superior to an OSFA approach, since SVANN is capable of both training model weights and selecting remote sensing indices as a function of location, whereas the rule-based approach consists of a static set of rules and input data for classifying wetlands all regions.

\subsubsection{SVANN Versus UNet OSFA}
To compare the SVANN models with the OSFA UNet model, we again made two sets of regional comparisons: In the Freshwater Forested/Shrub Wetland region, we compared SVANN Model 1 and the OSFA Model. In the Lakes Wetland region, we compared SVANN Model 4 and the OSFA Model. In the first comparison, we find that the OSFA UNet model performs slightly better than the SVANN model in precision, F1 score, and validation accuracy, with roughly equal recall. In the second comparison, the SVANN model has a slightly better recall, F1 score, and validation accuracy, with lower precision. Overall, the experimental results show that the SVANN models have roughly similar performance with the OSFA UNet model.

From these results, we find that using SVANN models to select remote sensing indices based on location achieves similar performance to an OSFA UNet model, even though the OSFA model uses roughly twice as much input data as well as an additional index feature during training. The findings support the claim that considering spatial heterogeneity using SVANNs during feature selection offers significant value, since only the most useful remote sensing indices are selected by the SVANN approach.

\begin{table*}
\caption{Experimental results comparing SVANN approach with OSFA.}
\label{tab:results_comparison}
\begin{tabular}{|l l l l l l l l l l l|}
\hline
\textbf{Model} & \textbf{Index} & \textbf{TN} & \textbf{FP} & \textbf{FN} & \textbf{TP} & \textbf{Precision} & \textbf{Recall} & \textbf{F1} & \textbf{Val. Acc.} & \textbf{Epochs} \\
\hline
\multicolumn{11}{|c|}{Test Area: Freshwater Forested/Shrub Wetland region}\\
SVANN Model 1 & NDVI & 695,391 & 522,235 & 255,792 & 2,983,030 & 0.851 & 0.921 & 0.885 & 0.827 & 8\\
SVANN Model 2 & NDWI & 874,312 & 672,671 & 249,550 & 2,659,915 & 0.798 & 0.914 & 0.852 & 0.817 & 8\\
\hline 
Rule-based Model 1 & NDVI & 1,066,717 & 480,266 & 403,777 & 2,505,688 & 0.839 & 0.861 & 0.850 & - & -\\
Rule-based Model 2 & NDWI & 1,115,080 & 431,903 & 518,482 & 2,390,983 & 0.847 & 0.822 & 0.834 & - & -\\
\hline
\textbf{OSFA Model} & \textbf{NDVI, NDWI} & \textbf{736,624} & \textbf{481,002} & \textbf{258,417} & \textbf{2,980,405} & \textbf{0.861} & \textbf{0.920} & \textbf{0.890} & 0.\textbf{831} & \textbf{8}\\
\hline
\hline
\multicolumn{11}{|c|}{Test Area: Lakes Wetland region}\\
SVANN Model 3 & NDVI & 2,074,210 & 1,039,475 & 405,994 & 1,526,593 & 0.595 & 0.790 & 0.679 & 0.737 & 8\\
\textbf{SVANN Model 4} & \textbf{NDWI} & \textbf{2,586,145} & \textbf{527,540} & \textbf{461,125} & \textbf{1,471,462} & \textbf{0.736} & \textbf{0.761} & \textbf{0.749} & \textbf{0.787} & \textbf{8}\\
\hline 
Rule-based Model 1 & NDVI & 2,313,277 & 800,408 & 467,235 & 1,465,352& 0.647 & 0.758 & 0.698 & - & -\\
Rule-based Model 2 & NDWI & 2,413,404 & 700,281 & 536,717 & 1,395,870& 0.666 & 0.722 & 0.693 & - & -\\
\hline
OSFA Model & NDVI, NDWI & 2,608,688 & 504,997 & 495,945 & 1,436,642 & 0.740 & 0.743 & 0.742 & 0.780 & 8\\
\hline
\end{tabular}
\end{table*}

\subsection{Interpretation}
{\color{black}
The experimental results of our wetland mapping case-study show the suitability of comparative physical interpretation of SVANNs using rule-based methods to achieve decomposability. 

For the test area having Forested/Shrub Wetland we find that SVANN Model 1 is better than Model 2 and for rule-based Model 1 is better than Model 2. If we interpret SVANN model 1 with rule-based Model 1, we can say that they have a higher chances that the weights correspond to the $NDVI > 0.1$, compared to that of SVANN Model 2. 

In contrast, if we compare the rule-based Model 1 with OSFA Model we cannot conlude whether the Model is more suitable for Forested/Shrub Area or Wetland Area. Similarly, for the test area having lakes Model 4 behaves similar to that of rule base model 2, which suggests that SVANN model 2 is more suitable for lakes.



We also discussed model interpretability in terms of simulatability, decomposability, and algorithmic transparency in Section \ref{sec:approach}. Based on the above interpretation we find that UNet based SVANNs are more decomposable than a OSFA approach, however limited in their simulatibility and algorithmic transparency due to the underlying use of UNet architecture. We believe to achieve additional transparency, SVANNs needs to be compared with more sophisticated yet highly interpretable models like partial differential equation (PDEs). Below, we initiate a discussion in that direction.}


\subsection{Multi Layer Neural Networks for PDE}
\label{subsec:PINN}
{\color{black} Representation learning allows multi-layer neural networks (MLNNs) to automatically derive features for a given set of inputs. In addition, by the universal approximator theorem \cite{cybenko1989approximation} a sufficiently large network can learn a significant range of functions. Thus, MLNNs can serve as a natural solver for PDEs. Use of MLNNs to solve PDEs has been explored as far back as the 1990s. For example, Lagaris et al. \cite{lagaris1998artificial} and Lee et al. \cite{lee1990neural} used neural networks to solve PDEs and ODEs on an a priori fixed mesh.

Recently, there has been a renewed interest in this direction due to significant advances in computing resources and software libraries such as TensorFlow \cite{abadi2016tensorflow}, which can efficiently compute high-dimensional derivatives using automatic differentiation \cite{baydin2018automatic}. For example, Raissi et al. \cite{raissi2017physics} constructed an estimate of PDEs using neural network models called Physics informed neural networks (PINNs). These networks can be considered as a class of data-efficient universal function approximators that naturally encode any underlying physical laws as prior information (e.g., through loss function). The approach allows for the estimation of physical models from limited data and leverages a priori knowledge that physical dynamics should obey a class of PDEs. More recent methods such as the Deep Galerkin Method \cite{sirignano2018dgm} do not assume such a mesh structure and are analogous to the Galerkin method, where the solution is approximated by neural networks, which can serve as a linear combination of basis functions. In this work, we limit our physical interpretation to PINNs and plan to explore additional interpretation(s) in the future work.}

For a better understanding of PINNs and automatic differentiation, we provide an example of solving a PDE using PINNs in Appendix \ref{App:PINN_EG} and details on automatic differentiation in Appendix \ref{App:AD}.

\subsection{Towards process-based model interpreta- tion using partial differential equations}
\label{subsec:PDE-based}
Partial differential equations (PDEs) have been studied extensively and are well-understood by various scientific communities. Therefore, representing PDEs with multi-layer neural networks (MLNNs) allows them to be simulatable by physicists, mathematicians, and other communities well-versed in PDEs. Further, partial differential equations are far more decomposable compared to MLNNs, as each term is a derivative, partial derivative, or a constant with well-defined properties. PDEs also offer increased transparency, as they can be dissected and studied mathematically and their error surface can be analyzed--manually for smaller equations, and through software for larger equations. 

Next, we give an example of geographically heterogeneous processes based on streamflow in flat and stepped rice fields. We model the process using SVANNs and the provide a physical interpretation using PINNs. 

\begin{figure}[htp]
\vspace{0pt}
  \centering
    \includegraphics[width=0.80\linewidth]{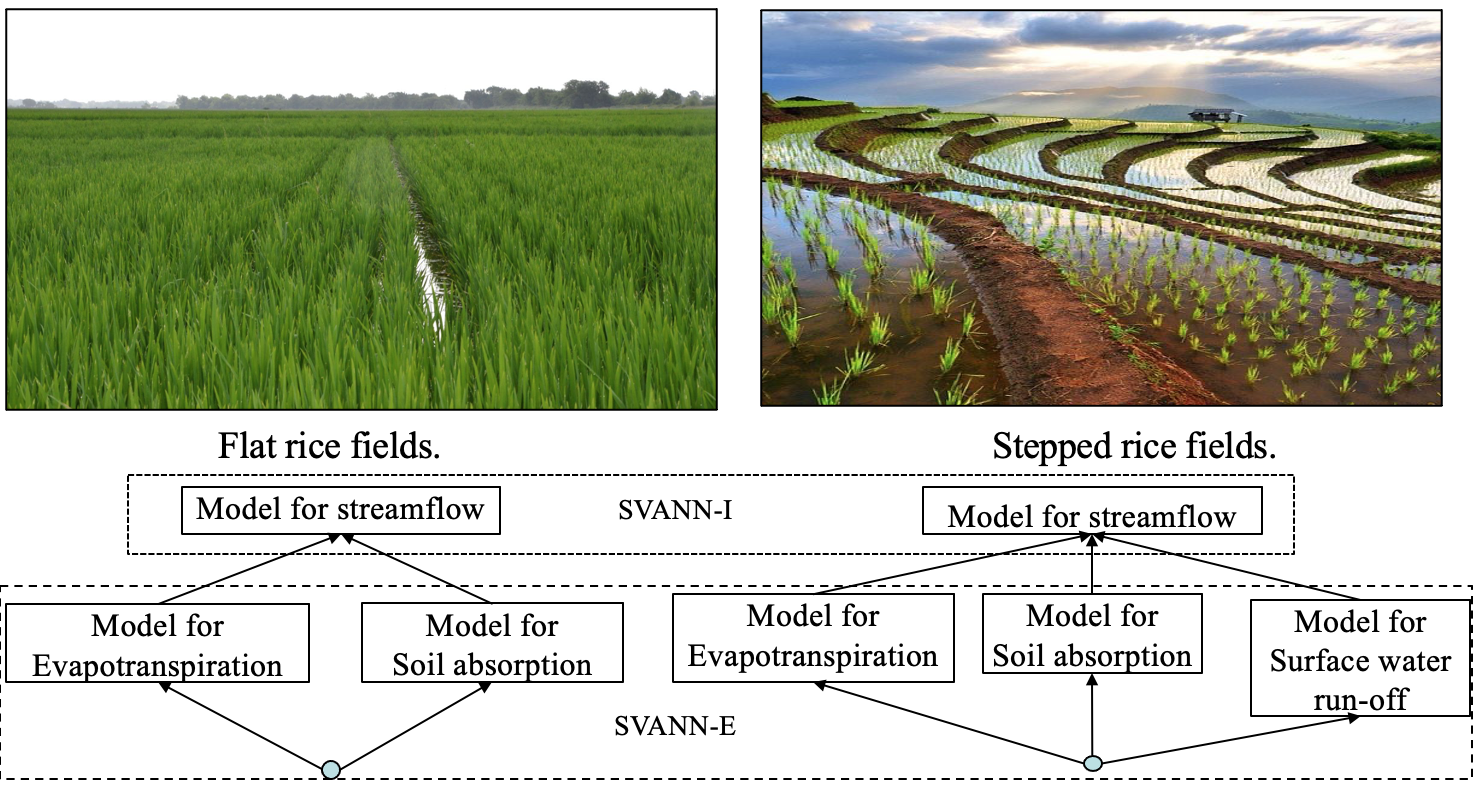}
  \caption{Geographically heterogeneous modeling of streamflow.}
  \label{fig:eg}
\end{figure}

Basic knowledge of Physics tells us that streamflow in rice paddies on the plains should differ from streamflow of rices paddies on non-planar step fields (Figure \ref{fig:eg}). Streamflow models therefore need different parameters (flat farm vs hillside agriculture) that reflect the different physical constraints across these regions.

Let $Q_i$ be the functional representation of surface water run-off for the $i^{th}$ zone and $lat^i$, $long^i$ be the sample coordinates within the $i^{th}$ zone. Then,
$$Q_1 = f_{flat}(lat^1, long^1)$$
$$Q_2 = f_{stepped}(lat^2, long^2)$$
where, let $Q_1$ denote the streamflow model for flat rice fields and $Q_2$ denote the streamflow model for stepped rice fields. For each functional representation, we define a partial differential equation (PDE)-based formulation as follows,
$$M_1 = \left( \frac{\partial^2 \hat{Q}_1}{\partial x^2} + 
            \frac{\partial \hat{Q}_1}{\partial x} - b \right)+
            [\hat{Q}_1^{(0)} - Q_1^{(0)}] + 
            [\hat{Q}_1^{(1)} - Q_1^{(1)}]$$
$$M_2 = \left( \frac{\partial^2 \hat{Q}_2}{\partial x^2} + 
            \frac{\partial \hat{Q}_2}{\partial x} - b \right)+
            [\hat{Q}_2^{(0)} - Q_2^{(0)}] + 
            [\hat{Q}_2^{(1)} - Q_2^{(1)}]$$
            
where $Q_i^{(i)}$ and $Q_i^{(1)}$ are the initial and boundary conditions for the function.

Let $Q_3$ represent a unified functional representation of the two models such that, 
$$Q_3 = f(Q_1, Q_2).$$ Then,
$$M_3 = \left( \frac{\partial^2 \hat{Q}_3}{\partial x^2} + 
            \frac{\partial \hat{Q}_3}{\partial x} - b \right)+
            [\hat{Q}_3^{(0)} - Q_3^{(0)}] + 
            [\hat{Q}_3^{(1)} - Q_3^{(1)}]$$
            
Also, let the average prediction error of a Model $M_i$ on a location (lat, long) be $Error_{Avg}(M_i, lat, long)$. Then we postulate,
$$ Error_{Avg}(M_3, lat^1, long^1) > Error_{Avg}(M_1, lat^1, long^1)$$
$$ Error_{Avg}(M_3, lat^2, long^2) > Error_{Avg}(M_2, lat^2, long^2)$$

Hence, using heterogeneous data to train models reduces their representational ability locally.

Further, SVANN-I approach will have similar structure for $f_{flat}$ and $f_{stepped}$, whereas, for SVANN-E approach the structure for $f_{flat}$ can differ from $f_{stepped}$ as shown in Figure \ref{fig:eg}.

Overall, we find that the models considered in this work had different physical interpretability. Table \ref{tab:mod_int} summarizes the interpretability of the models based on different interpretability metrics.
\begin{table}[htp]
\caption{Physical interpretability of the models considered.}
\label{tab:mod_int}
\resizebox{0.475\textwidth}{!}{%
\begin{tabular}{|c|c|c|c|}
\hline
\multirow{2}{*}{\textbf{Model}} & \multicolumn{3}{c|}{\textbf{Interpretability Metric}}                                  \\ \cline{2-4} 
                                & \textbf{Simulatibility} & \textbf{Decomposability} & \textbf{Algorithmic Transparency} \\ \hline
\textbf{OSFA (UNet)}  & Low    & Low    & Low  \\ \hline
\textbf{SVANN (UNet)} & Low & Medium & Low  \\ \hline
\textbf{Rule-based} & High   & High   & High \\ \hline
\textbf{PINN}        & High   & High   & High \\ \hline
\end{tabular}%
}
\end{table}

\section{Conclusion and future work}
\label{sec:CFW}
Given Spatial Variability Aware Neural Networks (SVANNs), the goal was to investigate mathematical (or computational) models for physical interpretation of SVANNs towards transparency. This problem is important due to several important use-cases (e.g., reusability, debugging, etc.) and challenging due to a large number of SVANN model parameters, vacuous bounds on generalization performance of neural networks, risk of overfitting, sensitivity to noise, etc. The related work on either model-specific or model-agnostic post-hoc interpretation is limited due to a lack of consideration of physical constraints (e.g., mass balance) and properties (e.g., second law of geography). To overcome the limitations of related work, we provide a physical interpretation of a SVANN using remote sensing indices which honor to spatial variability. Our results on the task of wetland mapping show that the proposed physical interpretations improve the transparency of SVANN models and the analytical results highlight the trade-off between model transparency and model performance (e.g., F1-score). We also describe an interpretation based on geographically heterogeneous processes modeled as partial differential equations (PDEs).


In the future, we plan to explore an experimental based validation of the PINN-based interpretation of SVANNs. In addition, we plan to expand our process-based interpretation of SVANNs to include additional neural network based solvers for PDEs. Finally, we plan to use an extensive set of remote sensing indices to further explore their use in the context of SVANNs and expand the validation to overlapping test areas.

\begin{acks}
This material is based on work supported by the National Science Foundation under Grant No. 1737633. We would also like to thank Kim Koffolt and the spatial computing research group for their helpful comments and refinements.
\end{acks}

\bibliographystyle{ACM-Reference-Format}
\bibliography{acmart}
\pagebreak
\appendix
\section{Precision, Recall, and F1-score}
\label{App:PRF}
Precision is a metric for prediction relevance. It calculates the percentage of the model's true predictions which were correctly classified. The formula for precision is given as:
\begin{equation}
    \text{Precision} = 1 - \text{Commission Error} = \frac{\text{TP}}{\text{TP + FP}}
    \label{eq: precision}
\end{equation}

Recall is a metric for prediction completeness. It calculates the fraction of total true predictions which the model correctly classified as true predictions. The formula for recall is given as:

\begin{equation}
    \text{Recall} = 1 - \text{Omission Error} = \frac{\text{TP}}{\text{TP + FN}}
    \label{eq: recall}
\end{equation}
There is inherently a trade-off between precision (i.e., prediction relevance) and recall (i.e., prediction completeness), which necessitates a metric which considers this interaction. As such, we used the F1 score, which is the harmonic mean of precision and recall, as another metric to evaluate the pixel-level classification performance (this metric is fundamentally similar to the 'dice coefficient' used by the computer vision community). While the precision and recall metrics can offer useful information on their own (i.e., prediction relevance and prediction completeness), the F1 score considers these two metrics as equally important in its calculation, which is given as:

\begin{equation}
    \text{F1 Score} = 2\cdot\frac{\text{Precision}\cdot\text{Recall}}{\text{Precision}+\text{Recall}}
    \label{eq: f1}
\end{equation}

With these three evaluation metrics, we obtain a robust baseline for quantitatively evaluating the ability of the model to correctly classify. Further, these metrics are commonly used when evaluating semantic segmentation models for wetland mapping tasks, which allows us to compare our results with rule-based approaches.

\section{Brief overview of UNet}
\label{App:UNET}
\begin{figure}[htp]
  \centering
  \includegraphics[width=0.9\linewidth, height=4.6cm]{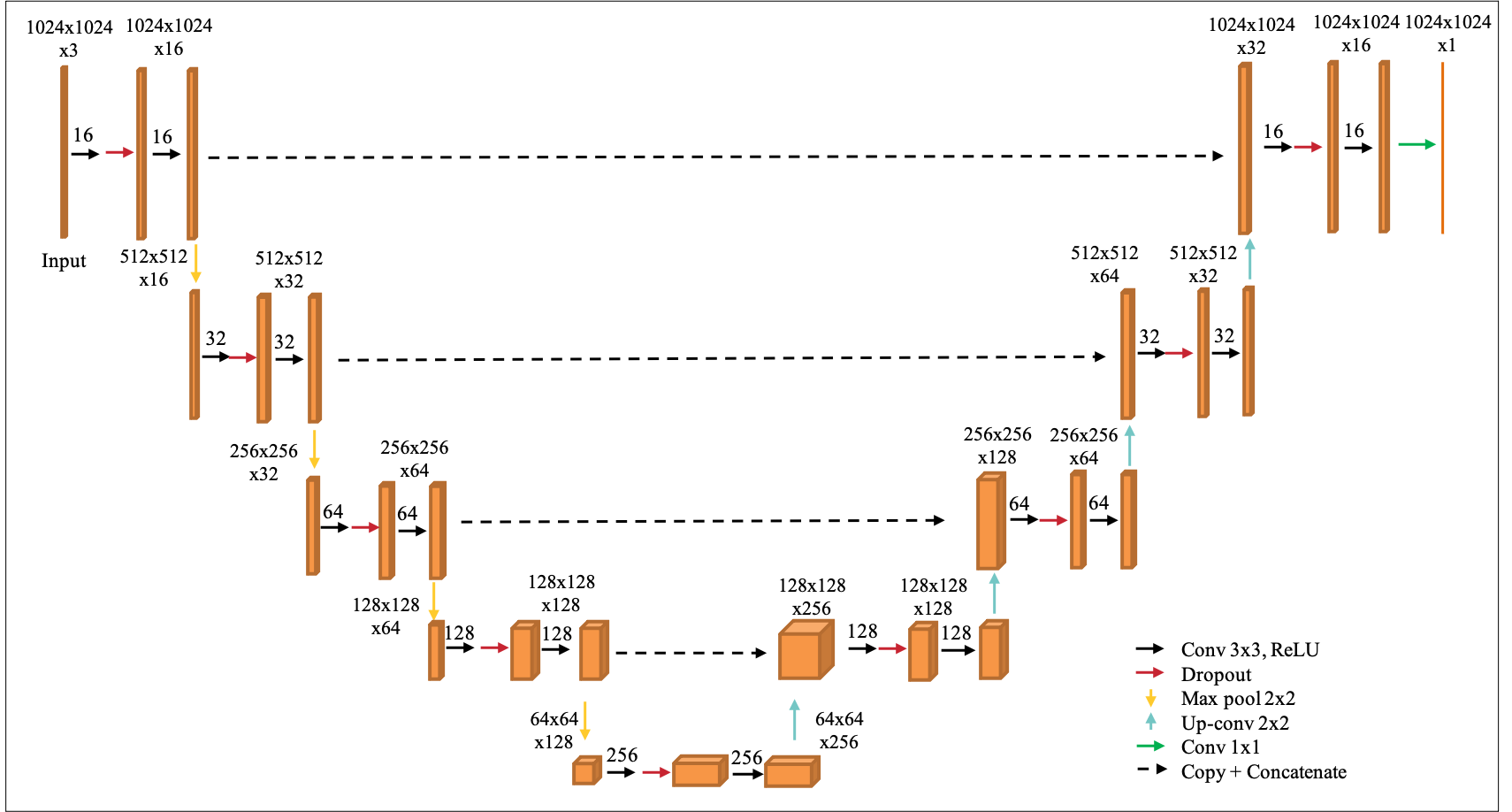}
  \caption{Architecture of UNet where arrows represent the operations (i.e., convolution, maxpool, dropout, and copy and concatenation). Feature dimensions are shown above the features and the number of channels in the convolution operation are shown above the arrows. (Best in color).}
  \label{Figure:UNet}
\end{figure}

Figure \ref{Figure:UNet} shows the UNet architecture adapted from its original paper \cite{ronneberger2015u}, presented earlier in \cite{gupta2020TIST}. UNet has two paths, a contracting path and an expanding path. Each step in the contracting path consists of two $3\times3$ convolutions (with padding), each followed by a rectified linear unit (ReLU) and a $2\times2$ maxpooling operation with a stride of $2$ for downsampling (which reduces each image dimension by half). Each step in the expanding path consists of up-sampling (i.e., up-convolution), which doubles the image dimensions, followed by a $2\times2$ convolution which reduces the number of feature channels by half. This is followed by concatenation with the corresponding feature map from the contracting path, and two $3\times3$ convolutions each followed by a ReLU. The final layer consists of a $1\times1$ convolution layer mapping each image pixel to the class. In addition, we use a dropout layer between each of the convolution layers to avoid over-fitting. Overall, UNet consists of 9 steps; 5 steps in the contracting path and 4 steps in the expanding path. 

\section{Example of Physics Informed Neural Networks to solve a PDE}
\label{App:PINN_EG}
We provide a toy example of solving PDEs using physics informed neural networks (PINNs). We consider a PDE defined as $\frac{\partial u}{\partial t} + 3.\frac{\partial u}{\partial x} = 0$ with initial condition as $u(x, 0) = x.e^{-x^2}$. 

\begin{figure}[htp]
\vspace{-10pt}
  \centering
     \includegraphics[width=0.75\linewidth]{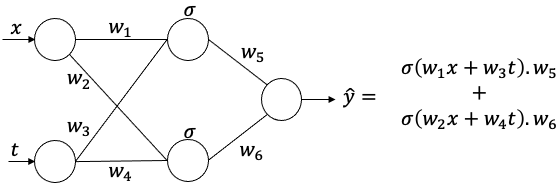}
  \caption{Architecture of the MLNN used in the toy example.}
  \label{Figure:MLNN-Arch}
\end{figure}

The PINNs model function $u(x,t)$ as $\hat{y}$, for which the loss function can be defined as a linear combination of the PDE and the initial condition as follows:
\\\textbf{Loss function} ($\mathcal{L}$) = $\left (\frac{\partial \hat{y}}{\partial t} + 3.\frac{\partial \hat{y}}{\partial x} \right ) + (\hat{y}(x, 0) - x.e^{-x^2})$
\\\textbf{Multi-layer neural network (MLNN) architecture:} For illustration, we use a basic MLNN with two input, one output, and one hidden layer, with 2 sigmoid nodes as shown in Figure \ref{Figure:MLNN-Arch}.  For simplicity, we use sigmoid nodes in the hidden layers and linear nodes in the input and output layer. Further, for the readability of equations we use the following two notations: First, $f_{w_iw_j} = \sigma(w_i.x + w_j.t) $ and Second, $f_{w_i} = \sigma(w_i.x)$. For example, the notations simplify $\hat{y}$ to $w_5.f_{w_1w_3} + w_6.f_{w_2w_4}$.
\\The loss function in terms of network parameters is as follows, 
\begin{multline}
  \mathcal{L} = w_5.f_{w_1w_3}.(1 + f_{w_1w_3}).(3.w_1 + w_3) \\ + w_6.f_{w_2w_4}.(1 + f_{w_2w_4}).(3.w_2 + w_4) \\ + f_{w_1}.w_5 + f_{w_2}.w_6 - x.e^{-x^2}.
\end{multline}

The weights of a given $x$ are updated using the back-propagation rule (Table \ref{Table:bp_wu}), where the loss function ($\mathcal{L}$) is used to update the weights as shown in Equations \ref{eq:dl_dw1}-\ref{eq:dl_dw6}. In theory, when the weights are updated, $\hat{y}$ should approach the theoretical value of the function represented by the PDE for a given input $x$. Table 3 shows the results for 5 iterations (i.e., loops) of $x=0.1$. It should be noted that the proposed architecture may not be the most appropriate for the given PDE, but simply illustrate the PINN approach.
\setlength{\belowdisplayskip}{0pt} \setlength{\belowdisplayshortskip}{0pt}
\setlength{\abovedisplayskip}{0pt} \setlength{\abovedisplayshortskip}{0pt}
\begin{table}[htp]
\small
\centering
\caption{Back-propagation weight update rule for the weights.}
\label{Table:bp_wu}
\begin{tabular}{|p{1.75cm} | p{1.75cm} | p{1.75cm} | p{1.75cm}| }
 \hline
 $\mathbf{w_1}$ & $\mathbf{w_2}$ & $\mathbf{w_3}$ & $\mathbf{w_4}$ \\ \hline
$w_1 + \eta.\frac{\partial \mathcal{L}}{\partial w_1}.x$ & $w_2 + \eta.\frac{\partial \mathcal{L}}{\partial w_2}.x$ & $w_3 + \eta.\frac{\partial \mathcal{L}}{\partial w_3}.x$ & $w_4 + \eta.\frac{\partial \mathcal{L}}{\partial w_4}.x$ \\ \hline 
 \multicolumn{2}{|c}{$\mathbf{w_5}$} & \multicolumn{2}{|c|}{$\mathbf{w_6}$} \\ \hline
 \multicolumn{2}{|c}{$w_5 + \eta.\frac{\partial \mathcal{L}}{\partial w_5}.\sigma(w_1.x + w_3.t)$} & \multicolumn{2}{|c|}{$w_6 + \eta.\frac{\partial \mathcal{L}}{\partial w_6}.\sigma(w_2.x + w_4.t)$} \\ \hline
\end{tabular}
\end{table}

\begin{multline}
\small
\frac{\partial \mathcal{L}}{\partial w_1} =  x.w_5.(3.w_1 + w_3).f_{w_1w_3}.(1 + f_{w_1w_3})^2 + w_5.f_{w_1w_3}.(1 + f_{w_1w_3}).\\(x.(3.w_1 + w_3).f_{w_1w_3} + 3) + x.w_5.f_{w_1}.(1 + f_{w_1})
\label{eq:dl_dw1}
\end{multline} 
\begin{multline}
\small
\frac{\partial \mathcal{L}}{\partial w_2} =  x.w_6.(3.w_2 + w_4).f_{w_2w_4}.(1 + f_{w_2w_4})^2 + w_6.f_{w_2w_4}.(1 + f_{w_2w_4}).\\(x.(3.w_2 + w_4).f_{w_2w_4} + 3) + x.w_6.f_{w_2}.(1 + f_{w_2})
\end{multline} 
\begin{multline}
\small
\frac{\partial \mathcal{L}}{\partial w_3} =  t.w_5.(3.w_1 + w_3).f_{w_1w_3}.(1 + f_{w_1w_3})^2 + w_5.f_{w_1w_3}.(1 + f_{w_1w_3}).\\(t.(3.w_1 + w_3).f_{w_1w_3} + 1)
\end{multline} 
\begin{multline}
\small
\frac{\partial \mathcal{L}}{\partial w_4} =  t.w_6.(3.w_2 + w_4).f_{w_2w_4}.(1 + f_{w_2w_4})^2 + w_6.f_{w_2w_4}.(1 + f_{w_2w_4}).\\(t.(3.w_2 + w_4).f_{w_2w_4} + 1)
\end{multline} 
\begin{equation}
\frac{\partial \mathcal{L}}{\partial w_5} = f_{w_1w_3}.(1 + f_{w_1w_3}).(3.w_1 + w_3) + f_{w_1}
\end{equation} 
\begin{equation}
\frac{\partial \mathcal{L}}{\partial w_6} = f_{w_2w_4}.(1 + f_{w_2w_4}).(3.w_2 + w_4) + f_{w_2}
\label{eq:dl_dw6}    
\end{equation}

\begin{table}[htp]
\small
\centering
\caption{Weight updates for the MLNN for $x=0.1, y=0.1$. The solution of the PDE at the sample point is $-0.208$.}
\label{tab:bp_weights}
\begin{tabular}{|p{0.5cm}|p{0.6cm}|p{0.6cm}|p{0.6cm}|p{0.6cm}|p{0.6cm}|p{0.6cm}|p{0.6cm}|p{0.5cm}|}
\hline
\textbf{Loop} & $\mathbf{w_1}$ & $\mathbf{w_2}$ & $\mathbf{w_3}$ & $\mathbf{w_4}$ & $\mathbf{w_5}$ & $\mathbf{w_6}$ & $\hat{y}$ & Loss\\
\hline
$0$ & $0.5$ & $0.5$ & $0.5$ & $0.5$ & $0.5$ & $0.5$ & $0.525$ & $2.01$\\
$1$ & $0.487$ & $0.487$ & $0.495$ & $0.495$ & $0.389$ & $0.389$ & $0.408$ & $1.52$\\
$2$ & $0.476$ & $0.476$ & $0.491$ & $0.491$ & $0.28$ & $0.28$ & $0.294$ & $1.05$\\
$3$ & $0.469$ & $0.469$ & $0.488$ & $0.488$ & $0.173$ & $0.173$ & $0.182$ & $0.6$\\
$4$ & $0.464$ & $0.464$ & $0.486$ & $0.487$ & $0.067$ & $0.067$ & $0.07$ & $0.17$\\
$5$ & $0.462$ & $0.462$ & $0.486$ & $0.486$ & $-0.038$ & $-0.038$ & $-0.04$ & $-0.25$\\
\hline
\end{tabular}
\end{table}

In this example, the equation is a transport equation whose general form is given by $\frac{\partial u}{\partial t} + v.\frac{\partial u}{\partial x} = 0$, where $v$ is a constant. The exact solution to the equation in the example is $u(x,t) = (x - 3t).e^{(x-3t)^2}$. Proving exact solution is out of the scope. 

\section{Automatic Differentiation}
\label{App:AD}
Automatic differentiation (AD) is based on the fact that all numerical computations are compositions of a finite set of elementary operations for which derivatives are known. Combining the derivatives of the constituent operations using the chain rule gives the overall composition derivative. In general, there are forward and reverse accumulating mode \cite{baydin2018automatic} of AD, but most of the mainstream machine learning libraries including TensorFlow and PyTorch implement the reverse accumulating mode, so we limit this discussion to the reverse accumulating mode. 

\begin{figure}[htp]
  \centering
  \includegraphics[width=0.8\linewidth]{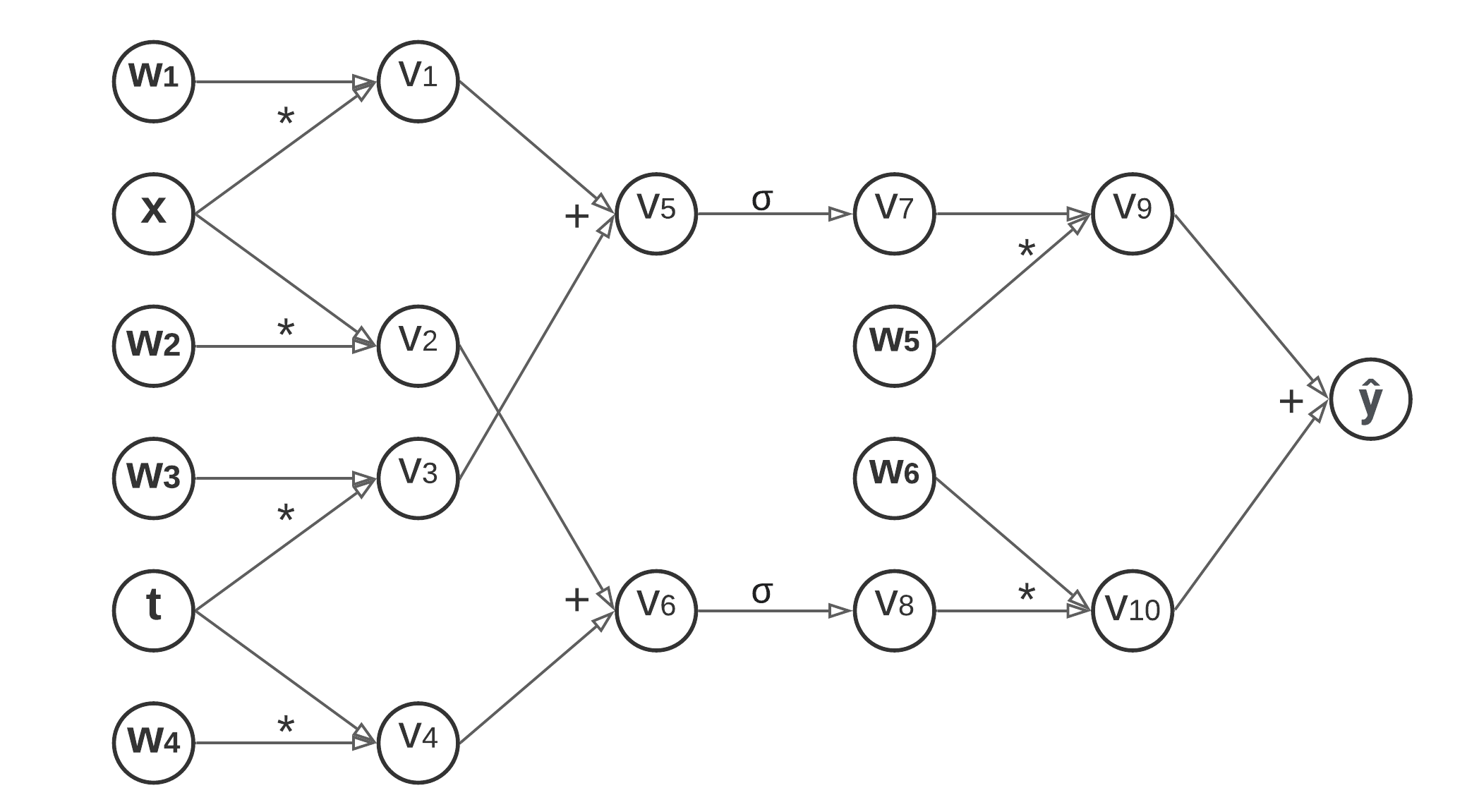}
  \caption{Computational graph of the toy example of MLNN described in Appendix \ref{App:PINN_EG}. All the known variables are marked in bold.} 
  \label{fig:comp_graph}
\end{figure}

\begin{table}[htp]
    \centering
    \caption{Numerical example of automatic differentiation using reverse accumulating mode. All the weights $w_i$ are initialized to $0.5$ for $ i=1,2,\dots,6$, and $x,t$ are initialized to 0.1. The order of computation to follow is from top to bottom for the forward pass, and bottom to top for the backward pass.}
    \begin{tabular}{|c|c|c||c|c|c|}
        \hline
        \multicolumn{3}{c}{Forward Pass} & \multicolumn{3}{c}{Backward Pass} \\
        \hline
        \hline
        Node & Eqa & Val & $\partial$ & Eqa & Val \\
        \hline
        $w_1$ & - & 0.5000 & $\frac{\partial \hat{y}}{\partial w_1}$ & $\frac{\partial \hat{y}}{\partial v_{1}} \frac{\partial v_{1}}{\partial w_1}$ & 0.0118\\
        $x$ & - & 0.1000 & $\frac{\partial \hat{y}}{\partial x}$ & $\frac{\partial \hat{y}}{\partial v_{2}} \frac{\partial v_{2}}{\partial x} + \frac{\partial \hat{y}}{\partial v_{1}} \frac{\partial v_{1}}{\partial x}$ & 0.1175\\
        $w_2$ & - & 0.5000 & $\frac{\partial \hat{y}}{\partial w_2}$ & $\frac{\partial \hat{y}}{\partial v_{2}} \frac{\partial v_{2}}{\partial w_2}$ & 0.0118\\
        $w_3$ & - & 0.5000 & $\frac{\partial \hat{y}}{\partial w_3}$ & $\frac{\partial \hat{y}}{\partial v_{3}} \frac{\partial v_{3}}{\partial w_3}$ & 0.0118\\
        $t$ & - & 0.1000 & $\frac{\partial \hat{y}}{\partial t}$ & $\frac{\partial \hat{y}}{\partial v_{4}} \frac{\partial v_{4}}{\partial t} + \frac{\partial \hat{y}}{\partial v_{3}} \frac{\partial v_{3}}{\partial t}$ & 0.1175\\
        $w_4$ & - & 0.5000 & $\frac{\partial \hat{y}}{\partial w_4}$ & $\frac{\partial \hat{y}}{\partial v_{4}} \frac{\partial v_{4}}{\partial w_4}$ & 0.0118\\
        \hline
        $v_1$ & $w_1.x$ & 0.0500 & $\frac{\partial \hat{y}}{\partial v_1}$ & $\frac{\partial \hat{y}}{\partial v_{5}} \frac{\partial v_{5}}{\partial v_1}$ & 0.1175\\
        $v_2$ & $w_2.x$ & 0.0500 & $\frac{\partial \hat{y}}{\partial v_2}$ & $\frac{\partial \hat{y}}{\partial v_{6}} \frac{\partial v_{6}}{\partial v_2}$ & 0.1175\\
        $v_3$ & $w_3.t$ & 0.0500 & $\frac{\partial \hat{y}}{\partial v_3}$ & $\frac{\partial \hat{y}}{\partial v_{5}} \frac{\partial v_{5}}{\partial v_3}$ & 0.1175\\
        $v_4$ & $w_4.t$ & 0.0500 & $\frac{\partial \hat{y}}{\partial v_4}$ & $\frac{\partial \hat{y}}{\partial v_{6}} \frac{\partial v_{6}}{\partial v_4}$ & 0.1175\\ 
        \hline
        $v_5$ & $v_1 + v_3$ & 0.1000 & $\frac{\partial \hat{y}}{\partial v_5}$ & $\frac{\partial \hat{y}}{\partial v_{7}} \frac{\partial v_{7}}{\partial v_5}$ & 0.1175\\
        $v_6$ & $v_2 + v_4$ & 0.1000 & $\frac{\partial \hat{y}}{\partial v_6}$ & $\frac{\partial \hat{y}}{\partial v_{8}} \frac{\partial v_{8}}{\partial v_6}$ & 0.1175\\
        \hline
        $v_7$ & $\sigma(v_5)$ & 0.5250 & $\frac{\partial \hat{y}}{\partial v_7}$ & $\frac{\partial \hat{y}}{\partial v_{9}} \frac{\partial v_{9}}{\partial v_7}$ & 0.5000\\
        $w_5$ & - & 0.5000 & $\frac{\partial \hat{y}}{\partial w_5}$ & $\frac{\partial \hat{y}}{\partial v_{9}} \frac{\partial v_{9}}{\partial w_5}$ & 0.5250\\
        $w_6$ & - & 0.5000 & $\frac{\partial \hat{y}}{\partial w_6}$ & $\frac{\partial \hat{y}}{\partial v_{10}} \frac{\partial v_{10}}{\partial w_6}$ & 0.5250\\
        $v_8$ & $\sigma(v_6)$ & 0.5250 & $\frac{\partial \hat{y}}{\partial v_8}$ & $\frac{\partial \hat{y}}{\partial v_{10}} \frac{\partial v_{10}}{\partial v_8}$ & 0.5000  \\
        \hline
        $v_9$ & $v_7.w_5$ & 0.2625 & $\frac{\partial \hat{y}}{\partial v_{9}}$ & $\frac{\partial \hat{y}}{\partial \hat{y}} \frac{\partial \hat{y}}{\partial v_{9}}$ & 1.0000\\
        $v_{10}$ & $v_8.w_6$ & 0.2625 & $\frac{\partial \hat{y}}{\partial v_{10}}$ & $\frac{\partial \hat{y}}{\partial \hat{y}} \frac{\partial \hat{y}}{\partial v_{10}}$ & 1.0000 \\
        $\hat{y}$ & $v_9 + v_{10}$ & 0.5250 & $\frac{\partial \hat{y}}{\partial \hat{y}}$ & - & 1.0000\\
        \hline 
    \end{tabular}
    \label{tab:auto_diff}
\end{table}

The reverse accumulating mode consists of a forward pass and a backward pass. The forward pass will define the computational graph and compute the numerical values of all the nodes starting from the input to the output in a feed-forward manner, and these values will be stored in the memory for efficient computation. When the forward pass is completed, the graph is evaluated in a backward pass to compute the gradient of each node (i.e., the partial derivative of the output node with respect to all the other nodes). Figure \ref{fig:comp_graph} gives an example of the computational graph derived from the toy example in Appendix \ref{App:PINN_EG}, and Table \ref{tab:auto_diff} provides an numerical example of how AD works.

\section{Resources}
We used Python's geopandas, rasterio, PIL and numpy libraries for preprocessing the imagery and shapefiles. We used Keras, a high-level deep learning interface for TensorFlow, to implement UNet. Model training and evaluation was done on a 3.70GHz Intel Core i7-8700K CPU, NVIDIA GeForce GTX 1080Ti GPU, and 16GB 2400MHz DDR4 RAM. Python based implementation of SVANN is at the following link: https://github.com/jayantgupta/SVANN.
\end{document}